\documentclass{article}
\usepackage{ijcai24}

\usepackage{times}
\usepackage{soul}
\usepackage{url}
\usepackage[hidelinks]{hyperref}
\usepackage[utf8]{inputenc}
\usepackage[small]{caption}
\usepackage{graphicx}
\usepackage{amsmath}
\usepackage{amsthm}
\usepackage{booktabs}
\usepackage{xcolor}
\usepackage{algorithm}
\usepackage{algpseudocode}
\usepackage[switch]{lineno}

\usepackage{amsmath,amsfonts,bm}

\def\eqref#1{equation~\ref{#1}}

\def\1{\bm{1}}

\def\rmW{{\mathbf{W}}}

\def\vx{{\bm{x}}}

\def\eve{{e}}

\def\mA{{\bm{A}}}

\def\mF{{\bm{F}}}
\def\mG{{\bm{G}}}

\def\mM{{\bm{M}}}

\def\mP{{\bm{P}}}

\def\mW{{\bm{W}}}
\def\mX{{\bm{X}}}

\DeclareMathAlphabet{\mathsfit}{\encodingdefault}{\sfdefault}{m}{sl}
\SetMathAlphabet{\mathsfit}{bold}{\encodingdefault}{\sfdefault}{bx}{n}

\def\gD{{\mathcal{D}}}

\def\gG{{\mathcal{G}}}
\def\gH{{\mathcal{H}}}

\def\gL{{\mathcal{L}}}

\def\gN{{\mathcal{N}}}

\def\gR{{\mathcal{R}}}
\def\gS{{\mathcal{S}}}

\def\sP{{\mathbb{P}}}

\def\sR{{\mathbb{R}}}

\def\emX{{X}}

\newcommand{\R}{\mathbb{R}}

\newcommand{\normltwo}{L^2}

\DeclareMathOperator*{\argmin}{arg\,min}

\usepackage{multicol,multirow}
\usepackage{booktabs}
\usepackage{bbding}

\newtheorem{theorem}{Theorem}

\title{Federated Prompt Learning for Weather Foundation Models on Devices}

\author{
Shengchao Chen
\and
Guodong Long\and
Tao Shen\and
Jing Jiang\And
Chengqi Zhang
\\
\affiliations
Australian Artificial Intelligence Institute, FEIT, University of Technology Sydney\\
\emails
shengchao.chen.uts@gmail.com,
\{guodong.long, tao.shen, jing.jiang, chengqi.zhang\}@uts.edu.au
}

\begin{document}

\maketitle

\begin{abstract}
On-device intelligence for weather forecasting uses local deep learning models to analyze weather patterns without centralized cloud computing, holds significance for supporting human activates. Federated Learning is a promising solution for such forecasting by enabling collaborative model training without sharing raw data. However, it faces three main challenges that hinder its reliability: \textbf{(1)} data heterogeneity among devices due to geographic differences; \textbf{(2)} data homogeneity within individual devices and \textbf{(3)} communication overload from sending large model parameters for collaboration. To address these challenges, this paper propose \textbf{Fed}erated \textbf{P}rompt learning for Weather Foundation Models \textbf{o}n \textbf{D}evices (\textbf{\texttt{FedPoD}}), which enables devices to obtain highly customized models while maintaining communication efficiency. Concretely, our \textit{Adaptive Prompt Tuning} leverages lightweight prompts guide frozen foundation model to generate more precise predictions, also conducts prompt-based multi-level communication to encourage multi-source knowledge fusion and regulate optimization. Additionally, \textit{Dynamic Graph Modeling} constructs graphs from prompts, prioritizing collaborative training among devices with similar data distributions to against heterogeneity. Extensive experiments demonstrates \textbf{\texttt{FedPoD}} leads the performance among state-of-the-art baselines across various setting in real-world on-device weather forecasting datasets.
\end{abstract}

\section{Introduction}
Climate change has a profound impact on both natural ecosystems and human societies~\cite{karl2009global,kjellstrom2016heat}. It leads to higher temperatures, sea level changes and more frequent extreme weather events~\cite{hagemann2013climate}. As a result, precise weather forecasting is becoming increasingly important. Data from meteorological devices in various regions is vital. However, analyzing this data with deep learning through centralized cloud computing presents challenges such as network dependence and privacy concerns~\cite{chakraborty2020comprehensive}. First, sending large volumes of data to centralized system places a heavy burden on communication networks, which is impractical for low-resource weather devices. Second, data from sensitive locations is subject to privacy laws, restricting sharing across devices~\cite{chen2023foundation}. To address these issues, on-device intelligence for analyzing data directly on the devices is crucial, reduces the need for data transfers, protects privacy, and decreases reliance on networks.

Federated Learning (FL)~\cite{mcmahan2017communication} is a promising method for on-device intelligence that trains a uniform model collaboratively across multiple devices without exchanging data. However, the models often underperform due to statistical heterogeneity among clients and data homogeneity on within individual clients' data. Personalized FL (PFL) offers new insights by developing specialized models for each device, enabling tailored on-device intelligence~\cite{paulik2021federated}. Recent PFL methods have introduced various methods to improve personalization~\cite{chen2022personalized,tan2022federated,li2021fedbn}. Despite these advances, two significant challenges remain. First, there is often inadequate consideration of the impact of physical geographic location on local models. For example, devices on seashores and hilltops may collect different data types even if they are geographically close. Second, the substantial communication demands of large neural networks burden both clients and servers. Edge devices with limited resources may struggle to process the necessary updates for these complex models~\cite{xiong2023feddm}. Moreover, the transfer of entire model parameters hampers communication efficiency.

To tackle the above issues, this paper introduces \textbf{Fed}erated \textbf{P}rompt Learning for Weather Foundation Models \textbf{o}n \textbf{D}evices (\textbf{\texttt{FedPoD}}), which allows devices to obtains high customized models with efficient communication. \textbf{\texttt{FedPoD}} comprising two pivotal components: (1) Adaptive Prompt Tuning and (2) Dynamic Graph Modeling. Adaptive Prompt Tuning against data homogeneity and reduces communication load via updating local prompts based on the frozen foundation model (FM) to capture local information and guide FM to generate accurate prediction, coupled with multi-level communication. Additionally, \textbf{\texttt{FedPoD}} uses Dynamic Graph Modeling on the server to manage prompts from clients and to build multiple graphs dynamically, considering various perspectives. This process takes geographic features into account and promotes priority collaborative learning among clients with similar data, mitigating the effects of data heterogeneity. As shown in Table~\ref{T1}, using a pre-trained foundation model leads to fewer parameters and higher performance compared to starting from scratch with FedAvg~\cite{mcmahan2017communication}. Furthermore, \textbf{\texttt{FedPoD}} achieves the best results with the proposed adaptive prompt tuning and dynamic graph modeling.
\begin{table}[H]\small
    \centering
    \resizebox{.47\textwidth}{!}{
    \begin{tabular}{ccc}
    \toprule
        Method & Trainable Param. & MAE/RMSE\\
        Train from scratch (\texttt{FedAvg}) & 5,284,173 & 40.3/51.2\\
    \midrule
        Pre-trained FM (\texttt{FedAvg}) & 215,089 & 33.5/44.5\\
        Pre-trained FM $\&$ Prompts (\texttt{FedAvg}) & \textbf{159,649} & 31.1/41.9 \\
        \textbf{\texttt{FedPoD}} (Ours) & \textbf{159,649} & \textbf{27.0}/\textbf{37.6}\\
        \bottomrule
    \end{tabular}}
    \vspace{-8pt}
    \caption{Compared with training Encoder-only Transformer as the foundation model. Experiments are implemented with FedAvg, and our method. Communication rounds: 30, local updating: 5.}
    \label{T1}
    \vspace{-8pt}
\end{table}

\paragraph{Main Contributions.} With extensive experiments across datasets including real-world on-deivce weather series datasets on various setting, we show that our \textbf{\texttt{FedPoD}} consistently outperforms state-of-the-art baselines. Besides, we conduct further  analysis to provide more insights in \textbf{\texttt{FedPoD}} from the perspective of ablation, hyperparameter sensitivity, and privacy. The main contributions is as follows:
\begin{itemize}
    \item We present \textbf{\texttt{FedPoD}}, a communication-efficient framework for on-device weather forecasting that addresses the challenges of data heterogeneity among devices and data homogeneity within individual clients during federated learning.
    \item We show \textit{Adaptive Prompt Tuning} that uses prompts to represent information and guide generation. These prompts enable multi-level communication and knowledge sharing, reducing the impact of data homogeneity.
    \item We introduce \textit{Dynamic Graph Modeling} to create dynamic links between participants' prompts. This prioritizes collaborative optimization for clients with similar representations, enhancing personalization.
    \item  With extensive experiments, we show \textbf{\texttt{FedPoD}} consistently achieves the best and help improve the communication efficiency while keeping privacy, and adaptive prompt tuning also benefits baselines.
\end{itemize}

\section{Related Work}
\paragraph{Weather Forecasting.}
Weather forecasting is a crucial tool that analyzes the variations in weather patterns. Recently, weather forecasting has made significant strides by incorporating data-driven approaches~\cite{chen2011comparison,sapankevych2009time,voyant2012numerical}. RNNs have shown promising in weather forecasting~\cite{shi2015convolutional,grover2015deep}. Besides, Transformers~\cite{zhou2021informer,zhou2022fedformer,wu2021autoformer,chen2023tempee} can capture non-stationary changes, which have contributed to their widespread use in weather analysis. To overcome challenges caused by intricate spatial-temporal correlation, spatial-temporal modeling methods~\cite{yu2017spatio} can be an effective solution. However, these methods all focus on data-intensive centralized training, which poses a challenge to weather forecasting practices.

\paragraph{Personalized Federated Learning.}
Weather forecasting involves significant communication loads and raises privacy issues due to the large volume of data processes on parallel~\cite{chavan2017integrated}. Federated learning (FL)~\cite{mcmahan2017communication} offers a way to perform on-device intelligence but is often hampered by data heterogeneity and homogeneity. Personalized FL (PFL) seeks to overcome these issues by training customized models for each device, providing fresh insights. For example, \cite{t2020personalized,hanzely2020lower,li2021ditto} add a regularization that decomposes the personalized model optimization from the global. ~\cite{li2021fedbn,collins2021exploiting} share part of the model and keep personalized layers private. \cite{zhang2020personalized} enables a flexible method by adaptively weighted aggregation. \cite{fallah2020personalized} start from a Model-Agnostic Meta-Learning where a meta-model is learned to generate the initialized local model for each client. In addition, \cite{chen2022personalized} utilize structure information to explore the topological relations among clients.

\section{Preliminaries and Problem Formulation}
\paragraph{Weather Forecasting.}
A multivariate weather time series represented by $\mX_i \in \R^{m \times n}$, where $m$ and $n$ is the series length and the number of variables, respectively. Each data point is shown as $\vx_t \in \R^{1 \times n}$. The weather forecasting task can be divided into two categories below:
\begin{itemize}
    \item \textbf{Task 1-Multivariate to Univariate Forecasting}: Predicting a single variable for future $Q$ periods using all variables from the past $P$ periods.
    \item \textbf{Task 2-Multivariate to Multivariate Forecasting}: Predicting all variables for future $Q$ periods from all variables in the past $P$ periods.
\end{itemize}
These tasks can be defined as follows:
\begin{equation}\small
\begin{aligned} 
&\textbf{Task1:} \quad \left[ \vx_{t-P},  \vx_{t-P+1}, \cdots,  \vx_{t}\right] \stackrel{ f}{\longrightarrow}\left[ \vx_{t+1}^{T1},  \vx_{t+2}^{T1}, \cdots,  \vx_{t+Q}^{T1}\right], \\
&\textbf{Task2:} \quad \left[ \vx_{t-P},  \vx_{t-P+1}, \cdots,  \vx_{t}\right] \stackrel{ f}{\longrightarrow}\left[ \vx_{t+1}^{T2},  \vx_{t+2}^{T2}, \cdots,  \vx_{t+Q}^{T2}\right],
\end{aligned}
\end{equation}
where $f$ denotes the learning system, $ \vx_{t}^{T1} \in  \R^{1 \times 1}$ is the predicted variable at the $t$-th step, and $ \vx_{t}^{T2} \in  \R^{1 \times n}$ is the predicted variable at the $t$-th step.


\paragraph{On-device Weather Forecasting based on FL.}
Each device\footnote{We take ``device(s)`` and ``client(s)`` to mean the same one.} possesses a local data varying location pattern, leading to statistical heterogeneity. Thus, we can define the task on-device weather forecasting as:
\begin{equation}
    [f_1(D_1), f_2(D_2),..., f_N(D_N)] \rightarrow [D_1', D_2', ..., D_N']
\end{equation}
where the $D_k$ and $D_k'$ denote the input dataset and prediction in $k$-th client, respectively, and $f_k$ is the personalized model for $k$-th client. This makes vanilla FL that train an uniform model unsuitable, and the task is updated to the PFL problem that solves below bi-level optimization.
\begin{align}
    \notag F(v; w) \text{:} &= \argmin_{\lbrace v_1, v_2, ..., v_N \rbrace} \sum_{k=1}^{N} \frac{n_k}{n} F_k(v_k) + \lambda \gR(v_k, w), \\
    s.t. \quad &w \in \argmin_w G(F_1(w), F_2(w), ...,F_N(w)),
\end{align}
where each client hold a customized model parameterized by $v_i$, $w$ denotes the global model. $\gR(\cdot)$ is a regularization term, $G(\cdot)$ is the aggregation strategy. Previous studies have had difficulty managing the non.iid of geographic data, often overlooking how spatial-temporal correlation is affected by more than just location~\cite{chen2023prompt}. In this work, we aim to address two main challenges: \textbf{(1)} \textit{How can we ensure efficient communication between clients and servers while guaranteeing the framework's high performance?} \textbf{(2)} \textit{How can we minimize the heterogeneity caused by complex geographic features in the most cost-effective way?}

\section{Methodology}
\begin{figure*}
    \centering
    \includegraphics[width=.85\textwidth]{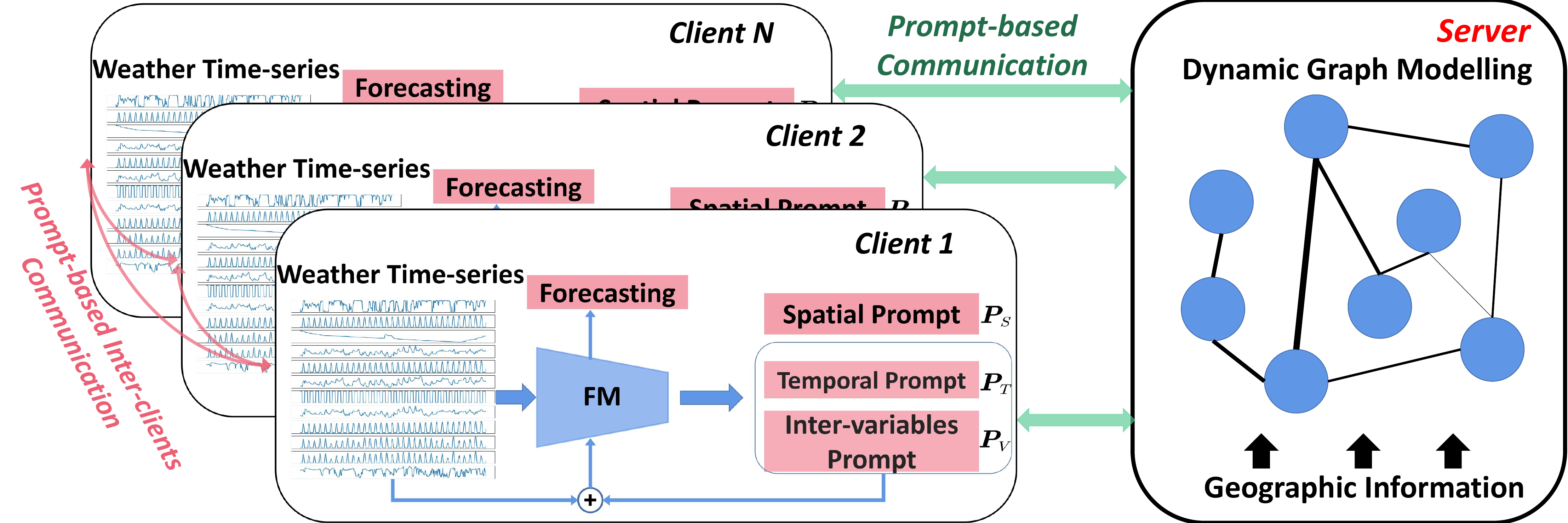}
    \caption{\small Architecture of \textbf{\texttt{FedPoD}}, prompts comprise the Spatial Prompt, Temporal Prompt, and Inter-variables Prompt. \textcolor{red}{$\leftrightarrow$}: communication exchanges prompts among clients, \textcolor{green}{$\leftrightarrow$}: communication between clients and the server only transmit prompts.}
    \label{fig:model_structure}
    \vspace{-8pt}
\end{figure*}

In this section, we detail our \textbf{\texttt{FedPoD}}, illustrated in Fig.~\ref{fig:model_structure}. Each client hold a pre-trained FM (PFM)\footnote{Detailed information about the utilized PFM in Appendix B.} and three types of prompts that updated locally. In each round, we introduce multi-level communication based on prompts uploaded by participants, including inter-clients and client-server. In the server, we present a novel aggregation method, Dynamic Graph Modeling, to building dynamic graphs based on structural information from prompts, reducing influence of data heterogeneity. With the updated prompts from the server, clients perform local optimization with a specialized prompt-wise loss. We'll describe them in more detail below.

\paragraph{Adaptive Prompt Tuning.} We introduce Adaptive Prompts Tuning for local update process to minimize the effects of data homogeneity within devices while keeping the computational load low. Unlike traditional prompt tuning in Natural Language Processing (NLP), which simply adjusts inputs to guide a pre-trained large language model (LLM) to produce desired outputs~\cite{white2023prompt}, our approach builds on this concept. It involves using lightweight prompts that dynamically represent local knowledge and act as  information carriers in multi-level communication. This helps lessen the overall impact of both global data heterogeneity and local data homogeneity during collaborative training. Specifically, we use trainable parameters as prompts, including \textsc{Temporal Prompts} ($\mP_T$) and \textsc{Inter-variables Prompts} ($\mP_V$), to capture the local time dynamics and the relationships among different variables. These prompts are incorporated into the time series and are refined during the local training phase. The updating process of $\mP_T$ and $\mP_V$ is shown in Alg.~\ref{alg1:localupdating}.

\begin{algorithm}[tbh]\small
  \begin{algorithmic}
  \State \textbf{Initialize} Original input series $\mX_{\text{ipt}}$, frozen PFM $\mF_M$, Temporal/Variable updating steps $K_t$ and $K_s$.
    \For{time forecasting step $q=1,2,...$}
    \State $\texttt{Updating}(\mF_M(\Vert  \mX_{\text{ipt}}, \mP_T\Vert^T)), \mP_T  \in \R^{q \cdot K_t \times n}$ \\ \Comment{$\Vert.\Vert^T$: \textit{concat} along temporal dimension}
    \State $\mP_T \leftarrow \Vert \mP_T, \mP'_T  \in \R^{K_t \times n}\Vert^T$ \\ \Comment{$\mP'_T$: Next temporal prompt block}
    \EndFor
    \For{variable forecasting step $p=1,2,...$}
    \State $\texttt{Updating}(\mF_M(\Vert  \mX_{\text{ipt}},  \mP_V\Vert^V)), \mP_V  \in \R^{m \times p \cdot K_v}$ \\ \Comment{$\Vert.\Vert^V$: \textit{concat} along variable dimension}
    \State $\mP_V \leftarrow \Vert \mP_V, \mP'_V  \in \R^{m \times K_v}\Vert^V$ \\ \Comment{$\mP'_V$: Next inter-variable prompt block}
    \EndFor
    \end{algorithmic}
    \caption{Implementation of $\mP_T$ and $\mP_V$ Updating}
    \label{alg1:localupdating}
\end{algorithm}

After updating the \textsc{Temporal Prompts} ($\mP_T$) and \textsc{Inter-variables Prompts} ($\mP_V$), we apply two learnable matrices, $\mW_{t}$ and $\mW_{v}$, to them. This is represented as $\mX = \mP_T \odot  \mW_{t} + \mP_V \odot \mW_{v}$. These matrices help to adjust the significance of the prompts, ensuring they contribute to our optimization goal without straying off course. Furthermore, we introduce \textsc{Spatial Prompt} ($\mP_S$), to encode local geographic pattern for comprehensive modeling, via updating with original input $\mX_{ipt}$ and $\mX$. The final prediction $\overline{\mX}$ is then derived using the following formula:
\begin{equation}
\begin{aligned}
    \mP_S, \mX \leftarrow &\texttt{LayerNorm}(\Vert \mX_{\text{ipt}}, \mX_{\text{geo}} \Vert, \Vert \mX, \mP_S \Vert ), \\
    &\overline{\mX} = \texttt{FFN}(\mF(\mX_{\text{ipt}} + \mX))
\end{aligned}
\label{eq:updatemps}
\end{equation}
where $\mX_{\text{geo}}$ denotes the client's geographic location represented by $(\phi, \lambda)$, $\phi$ and $\lambda$ is the latitude and longitude coordinates, respectively, for simultaneous updating of $\mP_S$ and $\mX$ to adjust the parameters of $\mP_S$.


\paragraph{Local Optimization from Multi-Task Perspective.} For each client's local optimization, we focus on two key elements: (1) multi-level communication regularization and (2) a multi-task perspective. The first involves interactions among clients and between clients and the server, aiming to mitigate the effects of data homogeneity. The second treats the optimization of various prompts as separate tasks, helping to lessen the unpredictability that comes with mixed updates. Consequently, we suggest a prompt-based optimization objective for local updates, as follows:
\begin{equation}
    \gL_{ap} = \texttt{MSE}(y', y) + \gR(\lbrace \mP_i \rbrace;\lbrace \mP_j \rbrace^l; \lbrace \mP_i \rbrace^l; \lbrace \mP \rbrace^*),
\end{equation}
where $\texttt{MSE}(\cdot)$ denotes the mean square error loss that evaluate the distance between ground-truth $y$ and output $y'$, $\gR(\lbrace \mP_i \rbrace;\lbrace \mP_j \rbrace^l; \lbrace \mP_i \rbrace^l;\lbrace \mP \rbrace^*)$ is the regularization term utilized to measure the distance between prompts, including personalized prompts $\lbrace \mP_i\rbrace^l $ of the $i$-th client, neighboring $j$-th client's prompts $\lbrace \mP_j \rbrace^l$, and global prompts $\lbrace \mP \rbrace^*$ obtained by averaging all client's prompts. The underlying motivations is allowing local client to craft highly customized models via decomposing prompt parameters from the neighboring and global prompts while keeping the comprehensive knowledge. Then, inspired by \cite{kendall2018multi}, we conceptualize the optimization from a multi-task view, can be formulated as:
\begin{equation}\small
    \begin{aligned}
    \gL_{ap} &= \texttt{MSE}(y', y) \\ &+\frac{1}{\xi^ 2} \normltwo(\lbrace \mP_i \rbrace, \lbrace \mP \rbrace^*) + \frac{1}{\xi^ 2} \normltwo(\lbrace \mP_i \rbrace, \lbrace \mP_i \rbrace^{l})\\
    &+ \frac{1}{\tau^2} \cdot \frac{1}{(|\gN| / S_{\mG}) - 1} \sum_{j\in \gN} \normltwo(\lbrace \mP_i \rbrace, \lbrace \mP_j \rbrace^l) \\ &+ 4 \lbrace \log_{2}(\xi) + \log_{2}(\tau) \rbrace.
    \end{aligned}
\label{LLF}
\end{equation}
Here, the $\xi$ and $\tau$ are importance coefficients that obey $\lambda, \tau \in (0, 1)$, the $\normltwo$ is L2 regularization (e.g., Euclidean distance, Cosine Similarity, etc.). $S_{\gG}$ represents the subgraph step used to adjust the scope of interaction between clients. The inter-client regularization term $\frac{1}{\tau^2} \cdot \frac{1}{(|\gN| / S_{\mG}) - 1} \sum_{j\in \gN} \normltwo(\lbrace \mP_i \rbrace, \lbrace \mP_j \rbrace)$, drives the local updating process towards a more comprehensive representation via considering neighboring clients with distinct feature distributions within a given range. Regulation terms $\frac{1}{\xi^ 2} \normltwo(\lbrace \mP_i \rbrace, \lbrace \mP \rbrace^*) $ and $\frac{1}{\xi^ 2} \normltwo(\lbrace \mP_i \rbrace, \lbrace \mP_i \rbrace^{l})$ are employed with the purpose of prompting the local clients to attain a more personalized representation.

\paragraph{Dynamic Graph Modeling for Global Aggregation.} We introduce Dynamic Graph Modeing (DGM) on the server to boost personalization by constructing spatial-temporal correlations among clients. This promotes collaborative optimization among clients with similar local knowledge representation. DGM uses the prompts shared by clients and their geographic data, like latitude and longitude, to form graphs. These graphs reveal possible relationships among clients, leading to a more customized optimization process. Specifically, we divide local prompts into three classes: (1) Temporal and Inter-Variables Prompts $\lbrace \mP_{T, i}, \mP_{V,i}\rbrace^N_{i=1}$; (2) Spatial Prompts $\lbrace \mP_{S,i} \rbrace^N_{i=1}$ and (3) Full Prompts $\lbrace \mP_i \rbrace^N_{i=1}$, where $N$ is the number of clients. First, the server generates a static graph $\mA_{\text{geo}}$ according to the location information based on Haversine formula~\cite{robusto1957cosine} as:
\begin{equation}\small
\begin{aligned}
2 R\cdot \tan^{-1}\bigg(\sqrt{\frac{\sin^2(\frac{\Delta\phi}{2}) + \cos(\phi_i) \cdot \cos(\phi_j) \cdot \sin^2(\frac{\Delta\lambda}{2})}{1-(\sin^2(\frac{\Delta\phi}{2}) + \cos(\phi_i) \cdot \cos(\phi_j) \cdot \sin^2(\frac{\Delta\lambda}{2})))}}\bigg),
\label{AGeoG}
\end{aligned}
\end{equation}
where $i,j \in \gN, i \neq j$, $\phi_i$ and $\phi_j$ are the latitude coordinates of client $i$ and $j$, respectively, $\Delta\phi = \phi_i - \phi_j$ is the difference in latitude between the two points in radians, $\Delta\lambda = \lambda_i - \lambda_j$ is the difference in longitude between the client $i$ and client $j$, $R$ is the radius of the Earth.

To grasp the potential correlations between clients dynamically, we use two matrices, $\mW_i$ are $\mW_j$, to apply linear transformations to the prompt vectors $\mP_i$ and $\mP_j$ of two different clients. The relation of the $i$-th client to the $j$-th client is calculated using the formula $\eve_{i,j} = \alpha (\mW_i \mP_i, \mW_j \mP_j)$, where $\alpha(\cdot)$ denotes a shared attention mechanism that operates in the space $\sR^{F'} \times \sR^{F'} \rightarrow \sR$. Here, $\rmW \in \sR^{F' \times F}$ helps determine the attention coefficients. We then introduce another matrix $\mW$ to calculate the weight of the connection (edge) and construct an adjacency matrix as follows:
\begin{equation}
\mA_{i,j} = \frac{e_{i,j}}{1 + e^{-\mW [\mW_i \mP_i -\mW_j \mP_j }]}.
\label{AG}
\end{equation}
For three types of prompts, we create three corresponding adjacency matrices (graphs), denoted as $\mA_{TV}$, $\mA_{S}$, and $\mA$, via Eq.~\ref{AG}. We then merge these with $\mA_{Geo}$ (from Eq.\ref{AGeoG}) using an attention mechanism to capture more precise correlation representations. Based on these matrices, we reconstruct prompts to deliver personalized prompts $ \lbrace P_i \rbrace^{l}$ for each client:
\begin{equation}
\begin{aligned}
    \mA' \leftarrow \texttt{Softmax}\left(\frac{(\mA_{Geo} - \mA_S)\mA_{TV}^\top}{\sqrt{d_k}}\right) \mA, \\
    \lbrace P_i \rbrace_{i=1}^{l, N} \leftarrow \alpha \mA \lbrace P_i \rbrace_{i=1}^N + (1 - \alpha) \mA' \lbrace P_i \rbrace_{i=1}^N,
\end{aligned}
\label{graphatt}
\end{equation}
where $\sqrt{d_k}$ is the dimension of adjacent matrix, and $\alpha$ is importance coefficient. The term $[\mA_{\text{Geo}} - \mA_{\text{S}}]$ highlights the discrepancy between the actual geographic correlation and the encoded spatial correlation, enabling the dynamic adjustment of spatial-temporal correlation among clients to achieve a more precise potential correlation graph modeling.

\begin{algorithm*}[tbh]
\caption{Implementation of \textbf{\texttt{FedPoD}}}\label{arg2}
\begin{algorithmic}[1]
\State \textbf{Initialize} local data $\lbrace D_i \rbrace^N_{i=1}$, foundation model $\mF_M$, prompts $\lbrace \mP_{T,i}, \mP_{V,i}, \mP_{S,i}\rbrace^N_{i=1}$
\State \textbf{Initialize} $\lbrace \mP_{T,i}, \mP_{V,i}, \mP_{S,i}\rbrace^N_{i=1}$ as $\lbrace \mP_i \rbrace^{N}_{i=1}$, $\mW_{bt, i}, \mW_{bs, i}$
\State \colorbox{red! 40}{\bf Server-side:} Broadcast frozen $\mF_{M}$ to each clients \Comment{Model communication}
\For{rounds $R=1,2,3...$} \Comment{FL rounds in sequence}
\State \colorbox{green! 40}{\bf Client-side:}
\State Download $\lbrace \mP\rbrace^l$ (\textcolor{red}{\it personalized prompts}), $\lbrace \mP\rbrace^*$ (\textcolor{red}{\it global prompts}) from the server
\For{each client $i$ in parallel} \Comment{Clients in parallel}
\State $\lbrace \mP_i\rbrace \leftarrow \textsc{LocalUpdate}(\mF_M, D_i, \lbrace \mP_i\rbrace^N_{i=1})$ \Comment{Prompt-based training}
\State Upload $\lbrace \mP_i\rbrace$ to the server \Comment{Model communication}
\EndFor
\State \colorbox{red! 40}{\bf Server-side:}
\State $\mA_{\text{geo}} \leftarrow \textsc{Haversine Formula}(\phi, \lambda)$ (\textcolor{red}{\it Eq.~\ref{AGeoG}}) \Comment{\textcolor{blue}{Generate the static graph}}
\State $\mA  \quad \leftarrow \textsc{Dynamic Graph Modeling}(\lbrace \mP_T \rbrace_i\rbrace^N_{i=1}, \lbrace \mP_V \rbrace_i\rbrace^N_{i=1}, \lbrace \mP_S \rbrace_i\rbrace^N_{i=1})$ (\textcolor{red}{\it Eq.~\ref{AG}}) \Comment{\textcolor{blue}{Generate the dynamic graph}}
\State $\mA_{\text{TV}} \leftarrow \textsc{Dynamic Graph Modeling}(\lbrace \mP_T \rbrace_i\rbrace^N_{i=1}, \lbrace \mP_V \rbrace_i\rbrace^N_{i=1})$ (\textcolor{red}{\it Eq.~\ref{AG}}) \Comment{\textcolor{blue}{Generate the dynamic graph}}
\State $\mA_{\text{S}}\;\ \leftarrow \textsc{Dynamic Graph Modeling}(\lbrace \mP_T \rbrace_i\rbrace^N_{i=1}, \lbrace \mP_V \rbrace_i\rbrace^N_{i=1})$ (\textcolor{red}{\it Eq.~\ref{AG}}) \Comment{\textcolor{blue}{Generate the dynamic graph}}
\State $\mA'\;\,\, \leftarrow \textsc{Attention}(\mA, \mA_{\text{TV}}, \mA_\text{S}, \mA_{\text{geo}})$ (\textcolor{red}{\it Eq.~\ref{graphatt}}) \Comment{\textcolor{blue}{Attention for filtering}}
\State $\lbrace \mP_i \rbrace_{i=1}^{l, N} \leftarrow \alpha \mA \lbrace \mP_i \rbrace_{i=1}^N + (1 - \alpha) \mA' \lbrace \mP_i \rbrace_{i=1}^N$ (\textcolor{red}{\it According to Eq.~\ref{graphatt}}) \Comment{\textcolor{blue}{Update personalized prompts}}
\State $\lbrace \mP_i\rbrace^* \leftarrow \frac{n}{n_k}\sum_{i=1}^{N} \mP^s$, $w_r \leftarrow \frac{n}{n_k}\sum_{i=1}^N w_{r, i} $ \Comment{\textcolor{blue}{Update global prompts and layers}}
\EndFor

\State \colorbox{black! 40}{\textbf{LocalUpdate}($\mF_M, D, \mP_\text{T}, \mP_\text{V}, \mP_\text{S}, \mF_{layer}$)}
\For{local epoch $e=1,2,...$}
\State Update $\mP_\text{T}, \mP_\text{V}$ (\textcolor{red}{\it Algorithm~\ref{alg1:localupdating}})
\State Update $\mP_\text{S}$ (\textcolor{red}{\it Eq.~\ref{eq:updatemps}})
\State Update rest of trainable parameters (\textcolor{red}{\it Eq.~\ref{eq:updatemps}}) 
\State Compute local loss (\textcolor{red}{\it Eq.~\ref{LLF}}) \Comment{Optimization from multi-task view}
\EndFor
\end{algorithmic}
\end{algorithm*}
\paragraph{Optimization for \textbf{\texttt{FedPoD}}.} The overall optimization objective of \textbf{\texttt{FedPoD}} is to solve a bi-level optimization problem, as below:
\begin{equation}\small
\begin{aligned}
     \argmin_{\lbrace \mP_i \rbrace;  \mA} \quad & \sum_{i=1}^{N} [\frac{n_i}{n} F_i(\lbrace \mP_i \rbrace; D_i) + \gR(\lbrace \mP_i \rbrace;\lbrace \mP_j \rbrace^l; \lbrace \mP_i \rbrace^l;\lbrace \mP \rbrace^*)]  \\ &+  \tau \mathcal{G} ( \mA),\\
     s.t. \quad & \lbrace \mP \rbrace^* \in \argmin_{\lbrace \mP_1 \rbrace,...,\lbrace \mP_N \rbrace} \sum_{i=1}^N \frac{n_i}{n}F_i(\lbrace \mP_i \rbrace), \\ &\lbrace \mP \rbrace^l \in \argmin_{\lbrace \mP_i \rbrace^l} \sum_{j \in \gN} \mA_{j,i} S(\lbrace \mP_i \rbrace^l, \lbrace \mP_j \rbrace^l)
\end{aligned}
\end{equation}
where $\lbrace \mP \rbrace$ denotes local prompts including $ \mP_T$, $ \mP_V$, and $ \mP_{S}$, $\lbrace \mP \rbrace^*$ is global prompts, the local model was parameterized by $\lbrace \mP \rbrace$ after receiving the pre-trained FM. The $\lbrace \mP_j \rbrace^l$ is personalized local models from other clients that achieve by the additional regularization term $\mathcal{G}(\cdot)$ that is a graph-based constraint that ensures each client aggregates with similar neighbor nodes. The learned graph with the adjacent matrix $\mA$ (computed by $\mA', \mA$) is expected to be sparse and able to preserve proximity relationships among clients. Algorithm~\ref{arg2} shows the detailed implementation of our \textbf{\texttt{FedPoD}}.

\section{Theorems and Proofs}
\begin{theorem}
Consider a on-device weather forecasting system with $m$ clients. Let $\gD_1, \gD_2, ..., \gD_m$ be the true data distribution and $\hat{\gD_1}, \hat{\gD_2}, ... , \hat{\gD_m}$ be the empirical data distribution. Denote the head $h$ as the hypothesis from $\gH$ and $d$ be the VC-dimension of $\gH$. The total number of samples over all clients is $N$. Then with probability at least $1-\delta$:
\begin{equation}\small
    \begin{aligned}
        &\max_{(\lbrace \mP_1 \rbrace, \lbrace \mP_2 \rbrace, ..., \lbrace \mP_m \rbrace)} \left| \sum_{i=1}^m \frac{|D_i|}{N}\gL_{ap, \gD_i} - \sum_{i=1}^m \frac{|D_i|}{N} \gL_{ap, \hat{\gD_i}} \right| \\
        &\leq \sqrt{\frac{N}{2} \log\frac{(m+1) |\lbrace\mP\rbrace|}{\delta}} + \sqrt{\frac{d}{N}\log\frac{eN}{d}}
    \end{aligned}
\end{equation}
\end{theorem}
\begin{theorem}[Transmitting Prompts Ensure Privacy]
Consider a device with a frozen pre-trained foundation model parameterized by $\theta_f$, and trainable prompts  parameterized by $\theta_p$ but initialized before updates. Transmitting these prompts can ensure privacy in multi-level communication.
\end{theorem}
\begin{proof}
    Detailed proofs can be found at Appendix C.
\end{proof}

\section{Experiments}
\label{others}
\paragraph{Datasets.}
Three weather multivariate time-series datasets from~\cite{chen2023prompt}, including AvePRE, SurTEMP, and SurUPS collected by 88, 525, and 238 devices, respectively. All three datasets cover the hour-by-hour variability of 12 weather-related variables, and detailed information can be found at Appendix A due to page limitation.
\begin{table*}[tbh]\small
  \centering
  \resizebox{.9\textwidth}{!}{
    \begin{tabular}{cccccccc}
    \toprule
    \multirow{2}[1]{*}{Fine-Tuning Strategy} &\multirow{2}[1]{*}{Method} & \multicolumn{2}{c}{\bf AvePRE} & \multicolumn{2}{c}{\bf SurTEMP} & \multicolumn{2}{c}{\bf SurUPS}  \\
    &  & \textbf{Task1} & \textbf{Task2} & \textbf{Task1} & \textbf{Task2} & \textbf{Task1} & \textbf{Task2} \\
    \midrule
    \multirow{8}[1]{*}{\bf Conventional Fine-tuning}&FedAvg~\cite{mcmahan2017communication}  & 34.6/44.8 & 56.0/90.1 &  47.6/64.4  &  56.5/78.3  & 53.5/74.2 & 54.1/74.6 \\
    &FedProx~\cite{li2020federated}  &  31.7/42.1  & 54.4/87.2 &  44.4/62.7  &  52.9/76.4  &  51.2/69.5 &  52.3/72.4  \\
    &Per-FedAvg~\cite{fallah2020personalized}  &  30.9/40.7  & 54.3/\underline{71.5} & 41.4/\underline{60.9} &  51.8/\underline{73.3}  &  50.2/69.7  & 51.7/71.8  \\
    &APFL~\cite{deng2020adaptive} & 32.5/43.8 & 56.1/84.9 &  46.2/63.1  &  59.4/77.3  & 54.3/73.7 & 53.8/73.4  \\
    &FedAMP~\cite{huang2021personalized} & 31.9/41.3 & 54.7/84.2 &  43.8/62.9  &  52.3/73.7 &   51.5/70.0 &  53.2/73.4   \\
    &FedATT~\cite{jiang2020decentralized} & 34.5/44.7 & 63.2/89.8 &  48.7/63.1  & 61.0/79.4   &  58.8/73.6  & 64.6/82/0  \\
    &pFedMe~\cite{t2020personalized}  & 32.2/42.7 & 64.0/85.2 &  42.9/61.8  &  \underline{50.7}/74.6  &  51.7/70.1  &  52.5/72.0  \\
    &SFL~\cite{chen2022personalized} &  \underline{30.0/40.2}  &  \underline{53.1}/81.2 &  \underline{39.9}/62.6  &   51.7/76.1 &  \underline{48.0/69.1}  &  \underline{51.0/70.4}  \\
    \midrule
    \multirow{9}[1]{*}{\bf Adaptive Prompt Tuning (Ours)}& FedAvg~\cite{mcmahan2017communication}  & 32.4/42.8 & 51.0/76.3 &  41.2/61.7  &  54.4/76.8  & 52.1/72.2 & 53.2/73.8 \\
    & FedProx~\cite{li2020federated} &  27.1/38.0  &  47.1/70.2 &  39.7/61.5  & 51.7/75.2   &  48.1/67.1  &  51.0/\underline{67.6} \\
    & Per-FedAvg~\cite{fallah2020personalized}  & 29.3/37.9   & 45.3/\underline{67.4}   &  \underline{37.8/60.0}  &  51.3/\underline{72.2}  &  \underline{47.6}/68.2  &  \underline{50.1}/69.5  \\
    & APFL~\cite{deng2020adaptive} & 29.5/38.7 & \underline{46.0}/67.7 &  38.6/64.2  &  55.7/75.7  & 56.2/67.1 & 59.7/68.2 \\
    & FedAMP~\cite{huang2021personalized} &  \underline{27.1/37.4} & 46.7/69.7 &  39.2/61.0  &  \underline{51.2}/73.1  &  51.5/67.9 &  52.1/69.3  \\
    & FedATT~\cite{jiang2020decentralized} & 30.5/40.8 & 58.7/79.7 &  38.4/63.7  &  52.4/79.1  &  50.9/70.0  &  53.5/72.6  \\
    & pFedMe~\cite{t2020personalized} & 28.2/39.7 & 47.5/69.9 &  38.5/61.4  &  \textbf{50.5}/74.1  &  48.4/\underline{66.9}  &  51.2/68.8  \\
     \cmidrule{2-8}
    & SFL~\cite{chen2022personalized} & 31.1/39.2   &  46.4/68.8  &  37.6/59.3  &  54.2/73.7  &  47.2/66.0  &  49.8/67.2  \\
    & \textbf{\texttt{FedPoD} (Ours)} & \textbf{23.7}/\textbf{32.9} & \textbf{44.3}/\textbf{65.5} &   \textbf{35.7}/\textbf{55.0} &  51.4/\textbf{71.2}  & \textbf{43.9/\textbf{62.5}} & \textbf{45.2}/\textbf{63.9}\\
    \midrule
    \multirow{2}[1]{*}{\bf Other Prompt Tuning} & PromptFL~\cite{guo2023promptfl} & 33.8/42.7 & 49.2/70.0 & 44.1/63.2 & 59.7/78.9 & 51.1/73.7 & 58.2/69.2\\
    & MetePFL~\cite{chen2023prompt} & 29.9/37.2 & 46.1/68.0 & 40.1/58.6 & 51.3/73.0 & 48.4/67.7 & 52.4/67.6 \\
    \bottomrule
    \end{tabular}}
    \vspace{-8pt}
    \caption{Main results with different local fine-tuning strategy (MAE/RMSE reported), including Conventional Fine-tuning and ours adaptive prompt tuning, a lower value means better performance. \textbf{Bold} and \underline{Underline} denote the best and second best respectively.}
  \label{performcom}
  \vspace{-6pt}
\end{table*}
\paragraph{Baselines.}
We compare with competitive FL/PFL methods, including FedAvg~\cite{mcmahan2017communication}, FedProx~\cite{li2020federated}, pFedMe~\cite{t2020personalized}, Per-FedAvg~\cite{fallah2020personalized}, FedATT~\cite{jiang2020decentralized}, APFL~\cite{deng2020adaptive}, FedAMP~\cite{huang2021personalized}, and SFL~\cite{chen2022personalized}, while keeping the local foundation model consistent. Details about baselines, the hyper-parameters of the used foundation model can be found at Appendix A. In addition, we adapt two fine-tuning methods for each baseline for evaluate our method's effectiveness, as below:
\begin{itemize}
    \item \textbf{Conventional Fine-tuning}: Update local FM with an FFN as the fine-tune head.
    \item \textbf{Adaptive Prompts Tuning (Ours)}: Update prompts with the frozen FM and multi-level communication.
    \item \textbf{Other Prompt Tuning}: Add parameters to input to updating  models~\cite{chen2023prompt,guo2023promptfl}.
\end{itemize}
\paragraph{Implementation.} The task of on-device weather forecasting is to predict the next 12 hours using the data from the previous 12 hours. 1. Main experiments are conducted in 25 local epoch within 50 federated communication round. Following~\cite{chen2022personalized}, Mean Absolute Error (MAE) and Root Mean Squared Error (RMSE) are used as evaluation metrics. All results are in $100\times$ the original value for a clearer comparison. Detailed information about the implementation for the FM, local updating process and the aggregation can be found at Appendix B due to page limitations.

\vspace{-2pt}

\begin{table*}[tbh]
    \centering
    \resizebox{.9\textwidth}{!}{
    \begin{tabular}{cccccccccc}
    \toprule
     Variant & $\mP_V$  & $\mP_T$ & $\mW_{bv}$ & $\mW_{bt}$ & $\mP_S$ & Federated Aggregation Strategy & Local Loss & \textbf{Task 1} & \textbf{Task 2}\\
    \midrule
     \multirow{2}[1]{*}{\bf \texttt{FedPoD-A}} & \multirow{2}[1]{*}{\textit{w/o}}  & \multirow{2}[1]{*}{\textit{w}} & \multirow{2}[1]{*}{-} & \multirow{2}[1]{*}{-} & \multirow{2}[1]{*}{\textit{w}} & \multirow{2}[1]{*}{$\lbrace P_i \rbrace_{i=1}^{l,N} \leftarrow \mA_\text{T} \lbrace P_i \rbrace_{i=1}^N + (1 - \alpha) \mA_\text{S} \lbrace P_i \rbrace_{i=1}^N$} & Ours & 29.9/40.4 & 53.7/78.4 \\
     &  &  &  &  &  &  & MSE & 31.7/42.4 & 54.4/80.0  \\
    \midrule
     \multirow{2}[1]{*}{\bf \texttt{FedPoD-B}} & \multirow{2}[1]{*}{\textit{w}}  & \multirow{2}[1]{*}{\textit{w/o}} & \multirow{2}[1]{*}{-} & \multirow{2}[1]{*}{-} & \multirow{2}[1]{*}{\textit{w}} &\multirow{2}[1]{*}{$\lbrace P_i \rbrace_{i=1}^{l,N} \leftarrow \mA_\text{S} \lbrace P_i \rbrace_{i=1}^N + (1 - \alpha) \mA_\text{V} \lbrace P_i \rbrace_{i=1}^N$}  &  Ours & \underline{28.2}/\underline{37.2} & 57.1/85.0 \\
      & &  &  &  &  &  & MSE & \underline{29.2}/\underline{39.0} & 58.2/85.9  \\
    \midrule
      \multirow{2}[1]{*}{\bf \texttt{FedPoD-C}} &  \multirow{2}[1]{*}{\textit{w/o}}  & \multirow{2}[1]{*}{\textit{w/o}} & \multirow{2}[1]{*}{-} & \multirow{2}[1]{*}{-} & \multirow{2}[1]{*}{\textit{w}} &\multirow{2}[1]{*}{$\lbrace P_i \rbrace_{i=1}^{l,N} \leftarrow \mA_\text{S} \lbrace P_i \rbrace_{i=1}^N$} &  Ours & 30.8/41.2 & 52.0/77.7 \\
       &  &  &  &  &  &  & MSE & 31.8/42.4 & 54.8/78.9  \\
    \midrule
      \multirow{2}[1]{*}{\bf \texttt{FedPoD-D}}  & \multirow{2}[1]{*}{\textit{w}}  & \multirow{2}[1]{*}{\textit{w}} & \multirow{2}[1]{*}{\textit{w}} & \multirow{2}[1]{*}{\textit{w}} & \multirow{2}[1]{*}{\textit{w/o}} &\multirow{2}[1]{*}{$\lbrace P_i \rbrace_{i=1}^{l,N} \leftarrow \mA_{\text{TV}} \lbrace P_i \rbrace_{i=1}^N$}  & Ours & 30.1/40.9 & \underline{48.7}/\underline{74.7} \\
      &   &  &  &  &  &  & MSE & 31.6/42.1 & \underline{50.9}/\underline{76.0}  \\
    \midrule
      \multirow{2}[1]{*}{\bf \texttt{FedPoD-D}} & \multirow{2}[1]{*}{\textit{w}}  & \multirow{2}[1]{*}{\textit{w/o}} & \multirow{2}[1]{*}{-} & \multirow{2}[1]{*}{-} & \multirow{2}[1]{*}{\textit{w/o}} &\multirow{2}[1]{*}{$\lbrace P_i \rbrace_{i=1}^{l,N} \leftarrow \mA_\text{V} \lbrace P_i \rbrace_{i=1}^N$} & Ours & 29.4/39.8 & 56.2/84.7 \\
      &   &  &  &  &  &  & MSE & 31.1/40.8 & 59.0/87.8 \\
    \midrule
      \multirow{2}[1]{*}{\bf \texttt{FedPoD-E}} &  \multirow{2}[1]{*}{\textit{w/o}}  & \multirow{2}[1]{*}{\textit{w}} & \multirow{2}[1]{*}{-} & \multirow{2}[1]{*}{-} & \multirow{2}[1]{*}{\textit{w/o}} & \multirow{2}[1]{*}{$\lbrace P_i \rbrace_{i=1}^{l,N} \leftarrow \mA_\text{T} \lbrace P_i \rbrace_{i=1}^N$} & Ours & 30.1/40.6 &53.7/79.0 \\
      &   &  &  &  &  &  & MSE & 31.7/43.5 & 54.2/80.5  \\
    \midrule
      \multirow{2}[1]{*}{\bf \texttt{FedPoD} (Ori.)} & \multirow{2}[1]{*}{\textit{w}}  & \multirow{2}[1]{*}{\textit{w}} & \multirow{2}[1]{*}{\textit{w}} & \multirow{2}[1]{*}{\textit{w}} & \multirow{2}[1]{*}{\textit{w}} &\multirow{2}[1]{*}{$\lbrace P_i \rbrace_{i=1}^{l,N} \leftarrow \alpha \mA \lbrace P_i \rbrace_{i=1}^N + (1 - \alpha) \mA' \lbrace P_i \rbrace_{i=1}^N$} & Ours & \textbf{23.7}/\textbf{32.9} & \textbf{44.3}/\textbf{65.5 }\\
      &  &  &  &  &  &  & MSE & \textbf{25.0}/\textbf{34.4} & \textbf{47.7}/\textbf{68.0}  \\
    \bottomrule
    \end{tabular}}
    \vspace{-8pt}
    \caption{Ablation results (MAE/RMSE report) about (1) \textit{Local Adaptive Prompts} and (2) \textit{Local Optimization Objective}, a lower value means better performance. \textbf{Bold}: the best, \underline{Underline}: the second best, \textit{w} and \textit{wo} denote the presence and absence of prompt, respectively. Note that $\mA_\text{T}$ and $\mA_\text{V}$ are generated by Eq.~\ref{AG} when either $\mP_T$ or $\mP_V$ is present alone.}
    \label{tab:abla}
    \vspace{-8pt}
\end{table*}

\subsection{Main Results}
Table~\ref{performcom} presents our main results, showing that our \textbf{\texttt{FedPoD}} outperforms baseline methods in most scenarios, often by a significant margin, across various tuning strategies. Notably, our \textit{adaptive prompt tuning} outperforms conventional fine-tuning while using about \textbf{74\%} parameters (see Table~\ref{T1}). This method enhances baseline models by enabling them to learn fewer parameters for considerable performance boosts. \textbf{\texttt{FedPoD}} records an average performance increase of \textbf{23.6\%/12.9\%}, \textbf{11.7\%/19.7\%}, and \textbf{12.6\%/4.3\%} over FedAvg, FedProx, and Per-FedAvg, respectively. These percentages reflect MAE improvements for Task1/Task2. The gains are particularly striking against SFL, which employs graph-based aggregation~\cite{chen2022personalized}. With adaptive prompt tuning, \textbf{\texttt{FedPoD}} improves by \textbf{9.8\%} and \textbf{6.7\%} on average. These figures rise to\textbf{15.9\%} for Task1 and \textbf{11.1\%} or Task2 with conventional fine-tuning. In addition, \textbf{\texttt{FedPoD}} can show a superior performance relative to related federated prompting methods, PromptFL~\cite{guo2023promptfl} and MetePFL~\cite{chen2023prompt}. We credit these benefits to two main strategies: (1) \textit{adaptive prompt tuning} guides the PFM to generate more accurate prediction based on lightweight prompts with multi-level communication, and (2) \textit{dynamic graph modeling} encourages collaborative optimization among clients with similarly distribution to mitigate data heterogeneity. These components effectively address data heterogeneity and homogeneity issues through a lightweight plug-and-play means.

\subsection{Framework Analysis}
\paragraph{Ablation Study.} We present the ablation study results from two angles: (1) examining local prompts and their aggregation method, and (2) assessing the local optimization objective. This helps confirm the effectiveness of our \textit{adaptive prompt tuning} and \textit{dynamic graph modeling} strategies. For (1), the findings in Table~\ref{tab:abla} reveal that: (i) ours multi-task local optimization objective outperforms the standard MSE in all ablation scenarios concerning prompts, and (ii) the lack of any kind of prompt significantly hinders performance due to inadequate local representation and global dynamic aggregation. Furthermore, the impact of our multi-task optimization objective is detailed in Table~4, with Term 1: $\frac{1}{\xi^ 2} \normltwo(\lbrace \mP_i \rbrace, \lbrace \mP \rbrace^*)$, Term 2: $\frac{1}{\xi^ 2} \normltwo(\lbrace \mP_i \rbrace, \lbrace \mP_i \rbrace^{l})$, Term 3: $\frac{1}{\tau^2} \cdot \frac{1}{(|\gN| / S_{\mG}) - 1} \sum_{j\in \gN} \normltwo(\lbrace \mP_i \rbrace, \lbrace \mP_j \rbrace^l)$. This indicates that omitting any single term of our local optimization objective leads to a drop in overall performance, underscoring the importance and necessity of each component.

\begin{table}[tbh]\small
  \centering
    \begin{tabular}{c|c|c|c|c}
    \toprule
     Term 1 & Term 2 & Term 3 & Task 1 & Task 2 \\
    \midrule
    \textit{w}   & \textit{wo}   & \textit{wo}   & 29.1/36.9 & 47.1/70.1 \\
    \textit{w}   & \textit{wo}   & \textit{w}   & \bf 27.3/36.3 & \bf 46.0/69.9 \\
    \textit{w}   & \textit{w}   & \textit{wo}   & 29.1/34.3 & \underline{46.6/72.5} \\
    \textit{wo}   & \textit{w}   & \textit{w}   & 29.0/34.6 & 47.9/74.8 \\
    \textit{wo}   & \textit{wo}   & \textit{w}   & \underline{28.2/37.0} & 49.2/74.4 \\
    \bottomrule
    \end{tabular}
  \label{tab:losstermc}%
  \vspace{-6pt}
    \caption{Ablation results about the multi-task optimization objective (MAE/RMSE report), a lower value means better performance. \textbf{Bold}: the best, \underline{Underline}: the second best.}
    \vspace{-8pt}
\end{table}%

\paragraph{Privacy.} We've implemented differential privacy (DP) in our \textbf{\texttt{FedPoD}} by adding random noise to the gradient updates. This noise is scaled by a factor of $\tau$, which we set to $1e^{-2}$. The impact of this addition is recorded in Table~\ref{tab:dp_exp}, which shows a drop in performance after incorporating DP. Despite this, as shown in Table~\ref{performcom}, \textbf{\texttt{FedPoD}} continues to surpass other baselines. Importantly, since \textbf{\texttt{FedPoD}} only uses adaptive prompts on the server to create graphs that capture the spatial-temporal relationships among clients, applying DP exclusively to these prompts is enough to maintain privacy.


\begin{table}[H]\small
  \centering
  \resizebox{.45\textwidth}{!}{
    \begin{tabular}{c|c|c|c|c}
    \toprule
    \multicolumn{2}{c|}{Method/Dataset} & \bf \texttt{FedPoD}  & \bf \texttt{FedPoD-DP} & Ave. Variation \\
    \midrule
    \multirow{2}[2]{*}{\textbf{AvePRE}} & Task1 & 23.7/32.9 & 24.8/33.9 & \textcolor{red}{$\downarrow$} 4.33\% \\
          & Task2 & 44.3/65.5 & 46.1/66.9 & \textcolor{red}{$\downarrow$} 2.88\% \\
    \midrule
    \multirow{2}[2]{*}{\textbf{SurTEMP}} & Task1 & 35.7/55.0 & 37.0/56.6 & \textcolor{red}{$\downarrow$} 2.69\% \\
          & Task2 & 51.4/71.2 & 52.7/73.0 & \textcolor{red}{$\downarrow$} 2.53\% \\
    \midrule
    \multirow{2}[2]{*}{\textbf{SurUPS}} & Task1 & 43.9/62.5 & 45.1/63.7 & \textcolor{red}{$\downarrow$} 2.33\% \\
          & Task2 & 45.2/63.0 & 46.4/65.2 & \textcolor{red}{$\downarrow$} 2.34\% \\
    \bottomrule
    \end{tabular}}
    \vspace{-6pt}
     \caption{Differential privacy experiment results (MAE/RMSE), \texttt{FedPoD-DP} denotes \texttt{FedPoD} with differential privacy.}
  \label{tab:dp_exp}
  \vspace{-8pt}
\end{table}%

\paragraph{Hyper-parameter Sensitivity.} We examine the impact of hyper-parameters from two angles: the \textit{prompt updating step} and the \textit{subgraph step}. Our configuration is as follows: we use 5 epochs for local updates and 10 communication rounds, while other settings follow our main experiments.  Table~\ref{tab:updatingstep} shows the results of the prompt updating step. The best performance for Task 2 is achieved with a step of 1, and for Task 1 with a step of 6. This inconsistency is due to the variable nature of weather patterns. Additionally, setting the step to 12 results in the poorest performance for both Task 1 and Task 2. This is because a single-step update does not account for the erratic periodicity of weather patterns, leading to inflexibility. Our findings on the impact of $\gS_G$ are shown in Table ~\ref{tab:subgraph}, where $\gS_G \in \lbrace 1, 2, 4, 6, 8, 10\rbrace$. The results suggests that \textbf{\texttt{FedPoD}} achieve the suboptimal when $\gS_G = 1$ across different tasks, while optimal results are achieved for Task 1 and Task 2 when $\gS_G = 10$ and $\gS_G = 2$, respectively. Bigger $\gS_G$ means more knowledge will be involved in local optimization. In our experiment, not all clients train in each round for training due to the considerable overhead. With a large $\gS_G$, clients are optimized locally with a restricted range of client knowledge, potentially overlooking valuable input from other participants and negatively affecting performance. We set $\gS_G = 1$ as the default because it considers all clients and allows for flexibility in specific scenarios. As only prompts $\mP_T, \mP_S,\mP_V$, which have fewer parameters, are involved, $\gS_G = 1$ does not significantly increase communication costs.
\begin{table}[H]\small
    \centering
    \vspace{-4pt}
    \resizebox{0.425\textwidth}{!}{
    \begin{tabular}{ccccc}
    \toprule
     Updating step of $\mP_T$  & Updating step of $\mP_V$ & Task & MAE & RMSE \\
     \midrule
     \multirow{2}[1]{*}{1} & \multirow{2}[1]{*}{1}  & \textbf{Task1} &39.9 & 50.2 \\
     &   & \textbf{Task2} & \textbf{51.5} & \textbf{79.5} \\
     \midrule
     \multirow{2}[1]{*}{2} & \multirow{2}[1]{*}{2}& \textbf{Task1} & 38.1 & 48.8 \\
     &   & \textbf{Task2} & 53.7 & 85.0 \\
     \midrule
     \multirow{2}[1]{*}{3} & \multirow{2}[1]{*}{3} & \textbf{Task1} & \underline{37.1} & 47.9 \\
     &   & \textbf{Task2} & 52.9 & 80.4 \\
     \midrule
     \multirow{2}[1]{*}{4} & \multirow{2}[1]{*}{4}  & \textbf{Task1} & 38.6 & \underline{47.7} \\
     &   & \textbf{Task1} & 53.2 & \underline{80.1} \\
     \midrule
     \multirow{2}[1]{*}{6} & \multirow{2}[1]{*}{6} & \textbf{Task1} & \textbf{35.7} & \textbf{46.1} \\
     &   & \textbf{Task2} & \underline{52.6} & 80.7 \\
     \midrule
     \multirow{2}[1]{*}{12} & \multirow{2}[1]{*}{12} & \textbf{Task1} & 39.3 & 50.3 \\
     &   & \textbf{Task2} & 53.7 & 84.8 \\
     \bottomrule
    \end{tabular}}
    \vspace{-8pt}
    \caption{Impact of prompt updating steps, \textbf{Bold}: the best, \underline{Underline}: the second best, a lower value means a better performance.}
    \label{tab:updatingstep}
    \vspace{-8pt}
    \end{table}
\begin{table}[H]\small
\centering
\vspace{-6pt}
\resizebox{0.3\textwidth}{!}{
    \begin{tabular}{cccc}
    \toprule
     Step of subgraph $\gS_G$ & Task & MAE & RMSE \\
     \midrule
     \multirow{2}[1]{*}{1}  & \textbf{Task1} & \underline{36.9} & \underline{47.0} \\
        & \textbf{Task2} & \underline{51.7} & 79.4 \\
    \midrule
     \multirow{2}[1]{*}{2}  & \textbf{Task1} & 39.1 & 49.9 \\
      & \textbf{Task2} &\textbf{51.5} & \textbf{79.0} \\
      \midrule
     \multirow{2}[1]{*}{4} & \textbf{Task1} & 38.1 & 49.7 \\
       & \textbf{Task2} & 51.8 & \underline{78.8} \\
     \midrule
     \multirow{2}[1]{*}{6}  &  \textbf{Task1} & 38.8 & 49.1 \\
        & \textbf{Task2} & 54.0 & 81.9 \\
     \midrule
     \multirow{2}[1]{*}{8}  & \textbf{Task1} & 39.3 & 49.7 \\
      & \textbf{Task2} & 54.4 & 81.8 \\
     \midrule
     \multirow{2}[1]{*}{10} & \textbf{Task1} & \textbf{35.9} & \textbf{45.6} \\
      & \textbf{Task2} & 52.4 & 79.6 \\
     \bottomrule
    \end{tabular}}
    \vspace{-8pt}
    \caption{Results about impact of subgraph step $\gS_G$, \textbf{Bold}: the best, \underline{Underline}: the second best. Lower means bette performance.}
    \label{tab:subgraph}
    \vspace{-4pt}
\end{table}

\section{Conclusion and Future Works}
In this paper, we seek to tackle the issue of data heterogeneity among devices and data homogeneity within individual devices in on-device intelligence weather forecasting. To achieve this, we propose a novel FL algorithm, \textbf{\texttt{FedPoD}}, which is built on two main components: Adaptive Prompt Tuning and Dynamic Graph Modeling. The former aims to mitigate data homogeneity via extracting latent knowledge with the frozen foundation model, alongside multi-level communication, and the last deals with data heterogeneity by prioritizing devices with similar distribution for aggregation and collaborative training based on prompt-related graphs. Extensive experiments on real-world on-device weather forecasting datasets shows \textbf{\texttt{FedPoD}} consistently outperforms state-of-the-art methods. However, \textbf{\texttt{FedPoD}} may struggle with predictions over very long periods due to the prompt updating. We plan to address this limitation in future research and expand our method to more on-device spatio-temporal prediction challenges.

\appendix

\section{Missing Information}
In this section, we supplement the missing information in the main text with related work, dataset details, and an introduction to baseline methods used in the experiments.

\subsection{Related Works}
\paragraph{Foundation Models}
Pre-trained foundation models (FMs) offer efficient scenario-specific solutions, leveraging their abundant parameters and data for understanding diverse downstream tasks with minimal data. Nowadays, pre-trained FMs have been proven great success in Natural Language Process~\cite{bommasani2021opportunities} and vision, such as ViT~\cite{dosovitskiy2020image}, Bert~\cite{devlin2018bert}, Dert~\cite{carion2020end}, and CLIP~\cite{radford2021learning}.Maximizing pre-trained model capability on low-resource devices gains attention across various real-world applications~\cite{chen2023prompt,tan2022federated}.

\paragraph{Prompt Learning.}
Prompt learning enhances language model efficiency~\cite{shin2020autoprompt,schick2020exploiting}, guiding relevant output via prompts. Due to its parameter efficiency and adaptability compared to fine-tuning, it's widely used in vision~\cite{yao2021cpt,zang2022unified,zhou2022towards,jia2022visual} and time-series~\cite{chen2023prompt,xue2022prompt}. Some works have introduce prompts to FL ~\cite{guo2022promptfl,zhao2022reduce,chen2023prompt,li2023visual} to reduce the computation cost~\cite{guo2022promptfl,zhao2022reduce} and achieve personalization~\cite{chen2023prompt,li2023visual}. However, these methods overlook the spatial-temporal correlation among clients with distinct geographical locations. Among them, Ref.~\cite{chen2023prompt} considers each variable as an individual node within a space and explores spatial associations between them rather than geographic location patterns. 

\subsection{Detailed Information about Dataset}
All three meteorological datasets based on multivariate time series proposed in our work are collected by \textbf{NASA} data website. The detailed information of these datasets in presented in Table~\ref{Dataset}.
\paragraph{AvePRE.} The dataset was collected by 88 meteorological satellites spanning a latitude and longitude range of \textbf{(38.41055, -91.08764)} to \textbf{(34.75988, -86.7999)}. The dataset contains 12 different meteorological variables designed for forecasting surface precipitation to prevent the negative impacts of extreme rainfall on human lives and properties. The dataset includes all data monitored by these sensing devices from \textbf{April 1, 2012,} to \textbf{February 28, 2016}.

\paragraph{SurTEMP.} The dataset was collected by 525 meteorological satellites and observatories spanning a latitude and longitude range of \textbf{(33.90689, 84.55078)} to \textbf{(30.63791, -79.56200)}. The dataset contains 12 different meteorological variables designed for forecasting surface temperature to prevent surface drought, which can cause sea level rise and ice melting. The dataset includes all data monitored by these devices from \textbf{January 3, 2019,} to \textbf{May 2, 2022}.

\paragraph{SurUPS.} The dataset was collected by 238 meteorological satellites, observatories, and solar radiation monitors spanning a latitude and longitude range of \textbf{(38.84179, 81.22352)} to \textbf{(37.03761, -76.90420)}. The dataset contains 12 different meteorological variables designed for forecasting upstream longwave flux to prevent regions from abnormal thunderstorm activity. The dataset includes all data monitored by these devices from \textbf{January 2, 2019,} to \textbf{July 29, 2022}.

All these datasets were observed in hours, where missing data beyond 12 consecutive hours are padded with zeros, while missing up to 2 consecutive hours are padded by interpolation.
\begin{table*}[tbh]\small
  \centering
     \resizebox{.65\textwidth}{!}{
    \begin{tabular}{cccc}
    \toprule
    Dataset & Period  & Devices & Features \\
    \midrule
    \multirow{12}[2]{*}{AvePRE} & \multirow{12}[2]{*}{\textbf{April 1, 2012} to \textbf{February 28, 2016}} & \multirow{12}[2]{*}{88} & Root zone soil wetness \\
       &    &    & Root zone soil moisture content \\
       &    &    & Mean land surface temperature \\
       &    &    & \textbf{Total surface precipitation} \\
       &    &    & Snow mass \\
       &    &    & Snow depth \\
       &    &    & Transpiration \\
       &    &    & Overland runoff \\
       &    &    & Fractional snow-covered area \\
       &    &    & Surface downward PAR beam flux \\
       &    &    & Evaporation from land \\
       &    &    & Total water stored in land reservoirs \\
    \midrule
    \multirow{12}[2]{*}{SurTEMP} & \multirow{12}[2]{*}{\textbf{January 3, 2019} to \textbf{May 2, 2022}} & \multirow{12}[2]{*}{525} & Long-wave radiation absorbed by the surface \\
       &    &    & Surface emissivity \\
       &    &    & Cloud area fraction in high cloud areas \\
       &    &    & \textbf{Surface temperature} \\
       &    &    & Surface albedo \\
       &    &    & Surface Incident Short Wave Stream \\
       &    &    & Optical thickness of all clouds \\
       &    &    & Surface downlink longwave traffic \\
       &    &    & Surface albedo for visible beam \\
       &    &    & Surface incident shortwave flux \\
       &    &    & Long-wave flux from the surface \\
       &    &    & Surface albedo of NIR beams \\
    \midrule
    \multirow{12}[2]{*}{SurUPS} & \multirow{12}[2]{*}{\textbf{January 2, 2019} to \textbf{July 29, 2022}} & \multirow{12}[2]{*}{238} & cloud area fractiom \\
       &    &    & Surface emissivity \\
       &    &    & Short-wave flux without aerosols \\
       &    &    & Surface temperature \\
       &    &    & Surface albedo \\
       &    &    & Flux of upwelling long waves \\
       &    &    & Short-wave flow \\
       &    &    & Downlink shortwave flux \\
       &    &    & \textbf{Downlink shortwave flux without aerosols} \\
       &    &    & Total upstream longwave flux \\
       &    &    & Short-wave flux \\
       &    &    & Rising flux without aerosols \\
    \bottomrule
    \end{tabular}}
      \caption{Information of three on-device weather forecasting datasets, which the \textbf{bold} is the forecasting weather-related variable in each dataset in the multivariate to unvariate forecasting task (\textbf{Task 1}).}
  \label{Dataset}%
\end{table*}

\subsection{Baselines}
We compare our proposed \textbf{\texttt{FedPoD}} with popular FL algorithms, including FedAvg, FedProx, pFedMe, PerFedAvg, FedATT, APFL, FedAMP, SFL, and MetePFL.

\paragraph{FedAvg.} Aggregating locally trained models to obtain a globally representative model via average strategy, while preserving the privacy of each individual's data.
\paragraph{FedProx.}An extension of FedAvg that adds a proximal term to the objective function to encourage closer alignment with the global model~\footnote{https://github.com/litian96/FedProx} .
\paragraph{pFedMe.} A pFL approach that adapts the global model to each user's local data distribution while taking into account the similarity between users to improve model generalization~\footnote{https://github.com/CharlieDinh/pFedMe}.
\paragraph{Per-FedAvg.} A variation of the FedAvg algorithm that allows for personalized model updates for each client by adding client-specific parameters to the global model and optimizing them in a decentralized manner during training~\footnote{https://github.com/ki-ljl/Per-FedAvg}.
\paragraph{FedATT.} An FL algorithm that uses attention techniques to address the heterogeneity of local data distributions (the code comes from this repository ~\footnote{https://github.com/dawenzi098/SFL-Structural-Federated-Learning}).
\paragraph{APFL.} A variant of Federated Learning that enables asynchronous communication among the clients, allowing them to perform local updates at their own pace and reducing the overall communication cost of the system (the code comes from this repository~\footnote{https://github.com/TsingZ0/PFL-Non-IID}).
\paragraph{FedAMP.} An FL algorithm that aims to improve the convergence speed and communication efficiency of federated optimization (the code comes from this repository ~\footnote{https://github.com/TsingZ0/PFL-Non-IID}).
\paragraph{SFL.} An PFL algorithm with graph structure information to make a more personalized model according to client-wise personalization~\footnote{https://github.com/dawenzi098/SFL-Structural-Federated-Learning}.
\paragraph{PromptFL.} An FL algorithm that employs trainable parameters to steer the local model towards generating more accurate outputs.
\paragraph{MetePFL.} A PFL algorithm for federated weather forecasting tasks integrates client-specific prompts into the local model for personalization.

\section{Foundation Model, Graphs, and Setting}
\subsection{Pre-trained Foundation Model}
\paragraph{Architecture}
The foundational model employed in this study is the Encode-only Transformer. Detailed information regarding the model's hyperparameter settings is presented in Table~\ref{tab:hper}.
\begin{table}[tbh]
    \centering
    \begin{tabular}{cc}
    \toprule
    Parameters / Strategy & Numbers \\
    \midrule
    Feature dimension  &  12\\
    Internal dimension of embeddings   &  256\\
    Number of heads & 8 \\
    Dimension of dense feedforward part & 256\\
    Dropout parameters & 0.3 \\
    Normalization & Group Norm \\
    Activation & ReLu\\
    Number of encoder layers & 4\\
    Position encoding & learnable\\
    \bottomrule
    \end{tabular}
    \caption{Hyper-parameters of the foundation model.}
    \label{tab:hper}
\end{table}

\paragraph{Pre-Training Strategy}
The pre-training strategy employed in our work for the Transformer foundation model on multivariate time series. In this approach, a binary noise mask, denoted by $\mM$, is independently created for each training sample and epoch, which is then applied to the input, denoted by $\mX$, resulting in the masked input $\hat{ \mX}= \mM \odot \mX$. For multivariate time series data, each variable is masked using an independent mask vector of length $w$, which alternates between segments of 0 and 1. The state transition probabilities are modeled as having a length that follows a geometric distribution with a mean of $l_m$. This is then followed by an unmasked segment of mean length $l_u = \frac{1-r}{r}l_m$, where $r$ is the masking probability. The mask rate $r$ in our work is set to 0.15, and mean masked length $l_m$ is set to 3. The objective function for the pre-training process is formulated as follows:
\begin{equation}
\mathcal{L}_{pre}=\frac{1}{mn} \sum_{i=1}^m\sum_{j=1}^n\left( \emX_{i,j}-\hat{ \emX}_{i,j}\right)^2,
\end{equation}
Here, $\emX$ and $\hat{\emX}$ represent the ground truth and forecasting value, respectively. However, the objective function differs from the MSE loss function in that it considers only the prediction values on the masked locations, instead of all the elements in the multivariate time series data. It is important to note that we perform FL-based pre-training, where the epoch of local training is set to 20 within a communication round of 20. The participation rate $C$ is 0.5, and the aggregation strategy is set to FedAvg~\cite{mcmahan2017communication} by default.

\subsection{Optimization of \textbf{\texttt{FedPoD}}}
\paragraph{Optimization for \textbf{\texttt{FedPoD}}.} The overall optimization objective of \textbf{\texttt{FedPoD}} is to solve a bi-level optimization problem, as below:
\begin{equation}\small
\begin{aligned}
     \argmin_{\lbrace \mP_i \rbrace;  \mA} \quad & \sum_{i=1}^{N} [\frac{n_i}{n} F_i(\lbrace \mP_i \rbrace; D_i) + \gR(\lbrace \mP_i \rbrace;\lbrace \mP_j \rbrace^l; \lbrace \mP_i \rbrace^l;\lbrace \mP \rbrace^*)]  \\ &+  \tau \mathcal{G} ( \mA),\\
     s.t. \quad & \lbrace \mP \rbrace^* \in \argmin_{\lbrace \mP_1 \rbrace,...,\lbrace \mP_N \rbrace} \sum_{i=1}^N \frac{n_i}{n}F_i(\lbrace \mP_i \rbrace), \\ &\lbrace \mP \rbrace^l \in \argmin_{\lbrace \mP_i \rbrace^l} \sum_{j \in \gN} \mA_{j,i} S(\lbrace \mP_i \rbrace^l, \lbrace \mP_j \rbrace^l).
\end{aligned}
\end{equation}
Here, $\lbrace \mP \rbrace$ represents local prompts, including $ \mP_T$, $ \mP_V$, and $ \mP_{S}$, while $\lbrace \mP \rbrace^*$ stands for global prompts. The personalized local prompts $\lbrace \mP_j \rbrace^l$ from neighboring clients are obtained through an additional regularization term $\mathcal{G}(\cdot)$, a graph-based constraint that ensures aggregation with similar neighboring nodes, and $S(I, J)$ is a distance measurements for the production of prompt matrices. The sparsity of the learned graph, represented by the adjacency matrix $\mA$ (calculated from $\mA', \mA$), is designed to maintain proximity relationships among clients. Specially, the function $S(I, J)$ can be formulated as:
\begin{equation}
    S[i][j] = \sqrt{\sum_{k=1}^{N} (p_{ik} - p_{jk})^2}
\end{equation}
where $S[i][j]$ is the distance between client $i$ and client $j$, $N$ denotes the dimension of each client's parameter, $p_{ik}$ denotes the $k$-th dimension parameter of client $i$'s prompts, and $p_{jk}$ denotes the $k$-th dimension parameter of client $j$'s prompts.

\subsection{Implementation}
The implementation of FedPoD is available at \textit{\url{https://github.com/shengchaochen82/FedPoD}}. In addition, we have distinct implementation for local updating process and the Dynamic Graph Modeling.
\paragraph{Setup of Local Updating.} The batch size is set to 256, and \textit{AdamW} with the weight decay $1e^{-4}$ and initial learning rate $1e^{-2}$ is adopted. The participant rate $C=0.3$ by default for three datasets, and the coefficients are $\gamma = 0.7$ and $\tau = 0.3$ (for Eq.~\ref{LLF}).

\paragraph{Setup of Dynamic Graph Modeling.} For the graph training, the epoch is 40 and the optimizer is SGD with learning rate of $1e^{-3}$, and the $\alpha = 0.99$ during aggregation.

\subsection{Evaluation Metrics}
Mean Absolute Error (MAE) and Root Mean Squared Error (RMSE) are utilized to evaluate the performance of our proposed FedWing and baseline, which can be formulated as
\begin{equation}
\begin{aligned}
     \text{MAE} = \frac{1}{nT} \sum_{i=1}^{n} \sum_{j=1}^{T} \left| y_{i,j} - \hat{y}_{i,j} \right|, \\
    \text{RMSE} = \sqrt{\frac{1}{nT} \sum_{i=1}^{n} \sum_{j=1}^{T} \left( y_{i,j} - \hat{y}_{i,j} \right)^2},
\end{aligned}
\end{equation}
where $n$ is the number of time series, $T$ is the number of forecasting periods, $y_{i,j}$ is the actual value of the $j$-th period of the $i$-th time series, and $\hat{y}_{i,j}$ is the predicted value of the $j$-th period of the $i$-th time series. Smaller MAE and RMSE means better model prediction performance.

\section{Theorems and Proofs}
\begin{theorem}
Consider a on-device weather forecasting system with $m$ clients. Let $\gD_1, \gD_2, ..., \gD_m$ be the true data distribution and $\hat{\gD_1}, \hat{\gD_2}, ... , \hat{\gD_m}$ be the empirical data distribution. Denote the head $h$ as the hypothesis from $\gH$ and $d$ be the VC-dimension of $\gH$. The total number of samples over all clients is $N$. Then with probability at least $1-\delta$:
\begin{equation}\small
    \begin{aligned}
        &\max_{(\lbrace \mP_1 \rbrace, \lbrace \mP_2 \rbrace, ..., \lbrace \mP_m \rbrace)} \left| \sum_{i=1}^m \frac{|D_i|}{N}\gL_{ap, \gD_i} - \sum_{i=1}^m \frac{|D_i|}{N} \gL_{ap, \hat{\gD_i}} \right| \\
        &\leq \sqrt{\frac{N}{2} \log\frac{(m+1) |\lbrace\mP\rbrace|}{\delta}} + \sqrt{\frac{d}{N}\log\frac{eN}{d}}.
    \end{aligned}
\end{equation}
\end{theorem}

\begin{proof}
    We start from the McDiarmid's inequality as
\begin{equation}
    \sP[g(X_1, ..., X_n) - \mathbb{E}[g(X_1, ..., X_n)] \geq \epsilon] \leq \exp{(-\frac{2\epsilon^2}{\sum_{i=1}^n c_i^2})}
\end{equation}
when
\begin{equation}
    \sup_{x_1, ..., x_n} |g(x_1,x_2,...,x_n) - g(x_1, x_2,...,x_n)| \leq c_i
\end{equation}
Eq.~15 equals to
\begin{equation}
    \sP[g(\cdot) - \mathbb{E}[g(\cdot)]\leq \epsilon] \geq 1 - \exp{(-\frac{2\epsilon^2}{\sum_{i=1}^n c_i^2})}
\end{equation}
which means that with probability at least $1 - \exp{(-\frac{2\epsilon^2}{\sum_{i=1}^n c_i^2})} $,
\begin{equation}
    g(\cdot) - \mathbb{E}[g(\cdot)] \leq \epsilon
\end{equation}
Let $\delta = \exp{(-\frac{2\epsilon^2}{\sum_{i=1}^n c_i^2})}$, the above can be rewritten as with the adaptive prompts at least $1 - \delta$,
\begin{equation}
    g(\cdot) - \mathbb{E}[g(\cdot)] \leq \sqrt{\frac{\sum_{i=1}^n c_i^2}{2} \log\frac{1}{\delta}}
\end{equation}
Now we substitute $g(\cdot)$ with our adaptive prompts as
\begin{equation}
    \max_{(\lbrace \mP_1 \rbrace, \lbrace \mP_2 \rbrace, ..., \lbrace \mP_m \rbrace)} \left( \sum_{i=1}^m \frac{|D_i|}{N}\gL_{ap, \gD_i} - \sum_{i=1}^m \frac{|D_i|}{N} \gL_{ap, \hat{\gD_i}} \right)
\end{equation}
we can obtain that with probability at least $1-\delta$, the following holds for specific adaptive prompts,
\begin{equation}\small
    \begin{aligned}
        & \max_{(\lbrace \mP_1 \rbrace, \lbrace \mP_2 \rbrace, ..., \lbrace \mP_m \rbrace)} \left( \sum_{i=1}^m \frac{|D_i|}{N}\gL_{ap, \gD_i} - \sum_{i=1}^m \frac{|D_i|}{N} \gL_{ap, \hat{\gD_i}} \right)\\
        & - \mathbb{E} \left[\max_{(\lbrace \mP_1 \rbrace, \lbrace \mP_2 \rbrace, ..., \lbrace \mP_m \rbrace)} \left( \sum_{i=1}^m \frac{|D_i|}{N}\gL_{ap, \gD_i} - \sum_{i=1}^m \frac{|D_i|}{N} \gL_{ap, \hat{\gD_i}}\right) \right] \\ 
        & \leq \sqrt{\frac{N}{2} \log \frac{1}{\delta}}
    \end{aligned}
\end{equation}
Considering there are $(m+1) |\lbrace \mP \rbrace|$ prompts in total ($\lbrace \mP \rbrace $ including $\mP_T, \mP_V$, $\mP_S$), by using Boole's inequality, with probability at least $1-\delta$, the following holds,
\begin{equation}\small
    \begin{aligned}
        & \max_{(\lbrace \mP_1 \rbrace, \lbrace \mP_2 \rbrace, ..., \lbrace \mP_m \rbrace)} \left( \sum_{i=1}^m \frac{|D_i|}{N}\gL_{ap, \gD_i} - \sum_{i=1}^m \frac{|D_i|}{N} \gL_{ap, \hat{\gD_i}} \right) \\
        & \leq \mathbb{E} \left[\max_{(\lbrace \mP_1 \rbrace, \lbrace \mP_2 \rbrace, ..., \lbrace \mP_m \rbrace)} \left( \sum_{i=1}^m \frac{|D_i|}{N}\gL_{ap, \gD_i} - \sum_{i=1}^m \frac{|D_i|}{N} \gL_{ap, \hat{\gD_i}} \right) \right] \\
        &+  \sqrt{\frac{N}{2} \log \frac{(m+1) |\lbrace \mP \rbrace|}{\delta}}
    \end{aligned}
\end{equation}
where $N$ is the total number of samples over all clients.
\begin{equation}
    \begin{aligned}
        &\mathbb{E} \left[\max_{(\lbrace \mP_1 \rbrace, \lbrace \mP_2 \rbrace, ..., \lbrace \mP_m \rbrace)} \left( \sum_{i=1}^m \frac{|D_i|}{N}\gL_{ap, \gD_i} - \sum_{i=1}^m \frac{|D_i|}{N} \gL_{ap, \hat{\gD_i}} \right) \right] \\
        &\leq \mathbb{E} \left[\sum_{i=1}^m \frac{|D_i|}{N} \max_{\lbrace \mP_i \rbrace}   \left(\gL_{ap, \gD_i} - \gL_{ap, \hat{\gD_i}} \right) \right] \\
        & \leq^{a} \sum_{i=1}^m \frac{|D_i|}{N} \mathcal{R} (\gH) \\
        & \leq \sum_{i=1}^m \frac{|D_i|}{N} \sqrt{\frac{d}{|D_i|} \log \frac{e|D_i|}{d}} \\
        & \leq \sum_{i=1}^m \frac{D_i}{N} \sqrt{\frac{d}{|D_i|} \log \frac{eN}{d}} \\
        &\leq^b \sqrt{\frac{d}{N}\log \frac{eN}{d}}
    \end{aligned}
\end{equation}
where $\gH$ is the hypothesis set of head $h$, $d$ is the VC-dimension of $\gH$. The $a$ follow from the definition of Rademacher complexity
\begin{equation}
    \mathcal{R}_n(\mathcal{F}) = \mathbb{E}_{\sigma}\left[\sup_{f\in\mathcal{F}}\frac{1}{n}\sum_{i=1}^{n}\sigma_i f(x_i)\right],
\end{equation}
where $\sigma_1, \sigma_2, \ldots, \sigma_n$ are independent Rademacher random variables that take values in $\lbrace -1, 1 \rbrace$ with equal probability,
$\mathbb{E}_{\sigma}$ denotes the expectation over the Rademacher variables,
$x_1, x_2, \ldots, x_n$ are the input data points, and the $b$ follows from Jensen's inequality, so
\begin{equation}
    \begin{aligned}
        & \max_{(\lbrace \mP_1 \rbrace, \lbrace \mP_2 \rbrace, ..., \lbrace \mP_m \rbrace)} \left| \sum_{i=1}^m \frac{|D_i|}{N}\gL_{ap, \gD_i} - \sum_{i=1}^m \frac{|D_i|}{N} \gL_{ap, \hat{\gD_i}} \right| \\
        & \leq \sqrt{\frac{N}{2} \log\frac{(m+1) |\lbrace\mP\rbrace|}{\delta}} + \sqrt{\frac{d}{N}\log\frac{eN}{d}}
    \end{aligned}
\end{equation}
\end{proof}

\begin{theorem}[Transmitting Prompts Ensure Privacy]
Consider a device with a frozen pre-trained foundation model parameterized by $\theta_f$, and trainable prompts  parameterized by $\theta_p$ but initialized before updates. Transmitting these prompts can ensure privacy in multi-level communication.
\end{theorem}
\begin{proof}
We assume that $f(x; \theta)$ is a convex function with respect to $\theta$, i.e., for any $\theta_1$ and $\theta_2$ and $\lambda \in [0, 1]$, we have
\begin{equation}
f(x; \lambda \theta_1 + (1 - \lambda) \theta_2) \leq \lambda f(x; \theta_1) + (1 - \lambda) f(x; \theta_2).
\end{equation}
Since only the local prompts $\lbrace \mP \rbrace$ ($\mP_T, \mP_V, \mP_S$) are shared between clients and the server, we redefine $\theta$ to $\theta' = [\theta_p', \theta_f]$, where $\theta_p'$ represents the updated prompt parameters from the server. We only modify $\theta_p$, leaving the pre-trained foundation model and the FFN-based head $\theta_f$ intact. Thus, data privacy can be ensured, as $\theta_f$ contains parameters that reveal device-specific information. Furthermore, the prompts are initialized randomly at the start of local training, adding a layer of uncertainty that helps protect device-specific information. Finally, the introduction of differential privacy (DP) techniques enhances privacy protection via adding random Gaussian noise to each client's gradient, effectively masking individual contributions and preventing the reconstrubution of sensitive information.
\end{proof}

\section{Other Discussion}
\subsection{Limitation}
Our experiments were performed on real datasets from hundreds of ground-based weather stations. However, many regions have thousands or tens of thousands of stations, and our current computational resources are insufficient to handle such large-scale scenarios. Additionally, our approach relies on the central server knowing the locations of each weather station, which might not be feasible in cross-country or global systems due to privacy protocols. Despite these limitations, our method holds considerable promise for global-scale weather forecasting on devices.

\subsection{Privacy}
FL poses a risk of data leakage, despite not sharing raw data among clients during the training of a shared model. An attacker might reconstruct the original data from the gradient updates sent by a client, particularly if the batch size and number of local training steps are small. This concern also applies in our framework when sharing prompt parameters ($\mP_V$, $\mP_T$, $\mP_S$) across clients. In our proposed framework, each participant maintains a private local model, which includes a frozen pre-trained foundation model, prompts, and an FFN-based head. The prompts that randomly initialized and head can be trained, but only the prompts are shared with the server and among clients. This selective sharing approach makes it harder to reverse-engineer the original data from gradients, as not all gradient information is disclosed. Furthermore, we add coefficients to the local loss function, complicating potential inference attacks aimed at deducing the original data, even as training continues indefinitely ($e \rightarrow \infty$).

\subsection{Additional Experiments}
We conducted an additional experiment to assess the sensitivity of the hyper-parameter $\alpha$ during aggregation. The results, presented below, are based solely on the AvePRE dataset with all other conditions consistent with previous experiments:
\begin{table}[tbh]
    \centering
    \begin{tabular}{ccc}
    \toprule
        Parameter & Task 1 & Task 2\\
    \midrule
        $\alpha=0.95$ &  24.4/35.2 & 46.3/66.6  \\
         $\alpha=0.96$ &  25.2/36.0 & \underline{45.5}/\underline{66.4}\\
        $\alpha=0.97$ & \underline{24.1}/36.2 & 46.0/67.4\\
        $\alpha=0.98$ & 26.2/\underline{33.1} & 46.1/66.5\\
        \bf $\alpha=0.99$ (Ori.) & \bf 23.7/32.9 & \bf 44.3/65.5 \\
    \bottomrule
    \end{tabular}
    \caption{Results on the effect of the hyper-parameter $\alpha$ during aggregation on the overall performance. \textbf{Bold}: the best, \textbf{Underline}: the second best.}
    \label{tab:my_label}
\end{table}

The results indicate that the initial configuration in our work, where $\alpha=0.99$, yields the best performance across various tasks. As we adjust the  
$\alpha$ from 0.95 to 0.98, the performance does not follow a consistent trend. This variability arises because changing the value of $\alpha$ affects the amount of information retained or filtered out during global aggregation, which is challenging to assess. Therefore, our choice is based on empirical evidence.

\subsection{Scalability for Large-scale Deployment}
The propose \textbf{\texttt{FedPoD}} can support large-scale deployment for the below reasons. (1) \textbf{\texttt{FedPoD}} eliminates the need for clients to train from scratch, reducing the demand for high computational power and large datasets, thus solving the communication efficiency issue of large-scale learning systems. (2) Devices update and transmit \textbf{$\textless 3\%$} of local parameters (refer to Table~\ref{T1}). These suggest that \textbf{\texttt{FedPoD}}  can be scaled
up to larger scenarios without incurring excessive costs. The
\textbf{\texttt{FedPoD}} has been validated on a system with 525 clients.

\subsection{Handling Extreme Weather Events}
Our \textbf{\texttt{FedPoD}} is evaluated on the weather datasets covering 9 years (2012-2016, 2019-2022) across 851 regions in total (Appendix A.2). In recent years, global warming and climate change impacted weather worldwide. We believe these datasets include some abnormal weather changes that might be defined as a kind of extreme weather event. Moreover, our proposed method leverages the graph relations among regions that can easily diffuse the detected extreme events from one region to other similar regions that could be nearby in geography or have similar weather patterns. 

\subsection{Physical Meaning of Prompts}
Here's the rationale and potential physical meaning behind our prompts: \textbf{Temporal Prompts:} Encapsulate the time-dependent aspects of weather, such as daily temperature cycles and seasonal changes, allowing the model to reveal and track temporal patterns in atmospheric conditions. \textbf{Inter-Variable Prompts:} delineate the interactions among meteorological variables, aiding the model in deciphering complex relationships like those between pressure systems, wind velocity, and rainfall events. \textbf{Spatial Prompts:} Reflect geographic and topographic influences on weather, capturing how local features like mountains or plains influence meteorological phenomena, enabling location-specific personalization.

\subsection{Trade-offs between Performance and Communication Overhead}
\textbf{\texttt{FedPoD}} tackles the performance-communication trade-off by updating \textbf{$\textless 3\%$} of local parameters, consisting exclusively of prompts-only prompts-achieving SOTA results. This approach cuts communication demands by transmitting minimal parameters instead of full local model, optimizing bandwidth and speeding up learning without sacrificing performance.

\clearpage
\bibliographystyle{named}
\bibliography{ijcai24}

\end{document}